\documentclass[letterpaper, 10 pt, conference]{ieeeconf}  

\IEEEoverridecommandlockouts                              

\usepackage{graphics} 
\usepackage{epsfig} 
\usepackage{amsmath} 
\usepackage{amssymb}  
\usepackage{color}
\usepackage{bm}
\usepackage{subfig}
\usepackage{cite}
    
\DeclareMathAlphabet\mathbfcal{OMS}{cmsy}{b}{n}

\title{\LARGE \bf
Active Estimation of 3D Lines in Spherical Coordinates* 
}

\author{Andr\'e Mateus, Omar Tahri, and Pedro Miraldo
\thanks{*This work was supported by FCT grant {\tt PD/BD/135015/2017} and project {\tt UID/EEA/50009/2019}. P. Miraldo was supported by the Swedish Foundation for Strategic Research (SSF), through the COIN project.}
\thanks{A. Mateus is with the Institute for Systems and Robotics, Instituto Superior T\'ecnico, Universidade de Lisboa, Portugal,
        EMail: {\tt\small andre.mateus@tecnico.ulisboa.pt}}%
\thanks{O. Tahri is with the INSA Centre Val de Loire, Universit\'e d'Orl\'eans, PRISME, Bourges, France,
        E-Mail: {\tt\small omar.tahri@insa-cvl.fr}}%
\thanks{P. Miraldo is with the Department of Automatic Control, KTH Royal Institute of Technology, Stockholm, Sweden,
        E-Mail: {\tt\small miraldo@kth.se}}%
}
\begin{document}

    \maketitle
\thispagestyle{empty}
\pagestyle{empty}

\begin{abstract}

Straight lines are common features in human made environments, which makes them a frequently explored feature for control applications.
Many control schemes, like Visual Servoing, require the 3D parameters of the features to be estimated.
In order to obtain the 3D structure of lines, a nonlinear observer is proposed.
However, to guarantee convergence, the dynamical system must be coupled with an algebraic equation.
This is achieved by using spherical coordinates to represent the line's  moment vector, and a change of basis, which allows to introduce the algebraic constraint directly on the system's dynamics.
Finally, a control law that attempts to optimize the convergence behavior of the observer is presented.
The approach is validated in simulation, and with a real robotic platform with a camera onboard.

\end{abstract}

\section{INTRODUCTION}
\label{sec:intro}

Straight lines are inevitable in any human made environment. 
Hence, they are a common feature used in applications like navigation, mapping, and control.
Compared with points, they provide more information about the world, and the one-to-one association is easier to obtain.
Besides, their detection in the image is also easier (e.g. Hough Transform \cite{matas2000}), and their tracking is more robust, as shown in \cite{rosten2005}.

In Visual Servoing one aims at controlling a camera to observe the selected features from a certain perspective point of view.
These features can be for instance points, image moments, and/or lines as shown in \cite{chaumette2006}. 
The apparent motion of the features w.r.t. the camera's velocity is modelled as a dynamical system.
Its state variables are the coordinates of the features in 2D \& 3D, the inputs are the camera's linear and angular velocities, and the output are the features projection to the image plane.
In \cite{andreff2002} this system is defined for 3D straight lines, represented in \emph{Pl\"ucker Coordinates} (see \cite{pottman2001} for details). 
These coordinates are the system's state, and the output is the normal vector to the plane containing the optical center of the camera, and the line.

Even though Visual Servoing is usually based on the features in the image plane (IBVS~\cite{chaumette2006}), this task depends on 3D information, namely point's depth and line's direction \& depth.
Thus, a method to recover the feature's 3D parameters should be coupled with Visual Servoing.
A typical solution to this problem is to consider consecutive camera frames, the displacement between them (assumed to be known), and use a filtering technique to handle noise and uncertainty, such as \cite{soatto1996,civera2008,civera2010,omari2013}.
Another approach is to use deterministic nonlinear state estimation methods, as shown in  \cite{deluca2008,corke2010,martinelli2012,dani2012}.
A framework for \emph{Active Structure-from-Motion} was proposed in \cite{spica2013}.
This framework consists on a nonlinear state observer and a control law that explores the system's linear velocity to achieve a desired convergence behavior.
It has been applied to points, cylinders \& spheres in \cite{spica2014}, planes in \cite{spica2015,spica2015b}, and rotational invariants in \cite{tahri2015}.
Recently, this framework was coupled with a Visual Servoing task in \cite{spica2017}.

Most of the previous approaches focused on points. 
In this work, however, the  goal is to use lines (Sec.~\ref{sec:prob}), and the observer in \cite{spica2013}. 
From the dynamics of a 3D line in a camera's reference frame in \cite{andreff2002} (Sec.~\ref{sec:lines_dyn}), one can conclude that the system is in the desired class of systems, after a simple coordinate change as shown in our previous work \cite{mateus2018} (Sec.~\ref{sec:obs_nonlinear}).
Nonetheless, the system does not verify the observability criterion of the framework.
In order to deal with this, the orthogonality constraint of the \emph{Pl\"ucker Coordinates} can be exploited, resulting in a Differential-Algebraic System (DAS).
Thus, the system in \cite{mateus2018} requires an algebraic equation to be solved after integrating the dynamics.
DAS are common in engineering, for instance in chemical, biological, and electrical systems. 
The design of observers for these systems, specially addressing nonlinear DAS can be found in \cite{aaslund2006,darouach2008,darouach2011,zheng2017}.

The output vector (measured in the image plane), which corresponds to the normal vector to the line's projection (interpretation plane \cite{hartley2003}), is transformed into spherical coordinates (Sec.~\ref{sec:spherical}).
Based on the fact that it is a unit vector, one can define an orthonormal basis with the output vector as one of the basis vectors (Sec.~\ref{sec:spherical_basis}).
To the remaining \emph{Pl\"ucker Coordinates} (not measured in the image), a change to the new basis is performed.
This yields a dynamical system (Sec.~\ref{sec:proposed_dyn}) comprised only of differential equations, with the orthogonality of the coordinates embedded in its dynamics.
Besides, the system is in the class of systems of \cite{spica2013}, and verifies its observability condition.
Then, an observer is designed for this system based on the framework in \cite{spica2013} (Sec.~\ref{sec:obs_active}). 
Besides a method for Active Vision is also presented.
Both the observer and the Active Vision approach are validated (Sec.~\ref{sec:results}), in simulation (Sec.~\ref{sec:simulation}), and with a real system comprised of a camera onboard a robot (Sec.~\ref{sec:exp_results}).
Finally, the conclusions are presented in Sec.~\ref{sec:conclusions}.

\section{Problem description}
\label{sec:prob}

This section presents the problem arising from the need to introduce an algebraic constraint to the system dynamics. 
It starts by defining the representation of straight lines using \emph{Pl\"ucker Coordinates} \cite{pottman2001}. 
Secondly, the dynamics of the apparent motion of a line in the image of a moving camera are presented.
Then, the class of systems in the scope of the observer proposed in \cite{spica2013} is introduced.
Finally, the violation of the observability criterion is identified.

\subsection{3D Straight Lines and their Dynamics}
\label{sec:lines_dyn}

Geometrically, 3D straight lines can be represented with four degrees of freedom as shown in \cite{roberts1988}.
However, to represent all 3D lines (including in infinity), more degrees of freedom are needed. 
One of those representations are the \emph{Pl\"ucker Coordinates} (see \cite{pottman2001} for more detail).

From \cite{andreff2002}, it is known that any 3D straight line can be explicitly represented using \emph{binormalized Pl\"ucker Coordinates}
\begin{equation}
    \mathbfcal{L} = \begin{bmatrix} \mathbf{d}\\ l\mathbf{h} \end{bmatrix},  \text{with} \quad \mathbfcal{L} \subset \mathcal{P}^5,
    \label{eq:biline}
\end{equation}
where $\mathcal{P}^5$ is the five-dimensional projective space; $\mathbf{d} \in \mathbb{R}^3$ is the line's direction vector with unit norm; $\mathbf{h} \in \mathbb{R}^3$ is the line's unit moment vector; and $l$ is the line's depth (i.e. the geometric distance between the line and the correspondent reference frame).
By considering the reference frame origin in the camera's optical center \cite{hartley2003}, the moment vector is the vector normal to the interpretation plane, which contains the line and the origin.
This vector is therefore defined by
\begin{equation}
    \mathbf{h} = \frac{\mathbf{p} \times \mathbf{d}}{|| \mathbf{p} ||\sin(\gamma)},
    \label{eq:h_def}
\end{equation}
where $\mathbf{p} \in \mathbb{R}^3$ is a point in the line, and $\gamma$ the angle between the point and the direction vector.
The line's  depth is  
\begin{equation}
    l = || \mathbf{p}|| \sin(\gamma),
\end{equation}
independently of $\mathbf{p}$.
From \eqref{eq:h_def} is trivial to conclude that
\begin{equation}
    \mathbf{h}^T\mathbf{d} = 0.
    \label{eq:ortho}
\end{equation}
The intersection  of the interpretation plane with the image plane yields the projection of the line. 
Since, the line's projection is also contained in the former, the moment vector can be measured directly from the image.

Let us consider a perspective camera looking at a straight line. 
Whenever the camera moves, the line's projection to image will also change\footnote{Unless the camera's optical center constantly belongs to the initial interpretation plane.}.
This apparent motion of the line in the camera's image can be model by a dynamical system. 
Its state are the \emph{Pl\"ucker Coordinates} of the line in \eqref{eq:biline}, its inputs are the camera's linear and angular velocity, and the output is the image measurement (the line's moment vector in this case).
The dynamics of this system are given by
\begin{align}
    \dot{\mathbf{d}} = &~ \bm{\omega}_c \times \mathbf{d}, \label{eq:ddyn} \\
    \dot{\mathbf{h}} = &~ \bm{\omega}_c \times \mathbf{h} - \frac{\bm{\nu}_c^T\mathbf{h}}{d}( \mathbf{d}\times \mathbf{h}) \label{eq:hdyn}, \ \text{and} \\
    \dot{l} = &~ \bm{\nu}_c^T(\mathbf{d} \times \mathbf{h}) \label{eq:depthdyn}\\
    \mathbf{y} = & \mathbf{h}, 
    \label{eq:y}
\end{align}
where $\bm{\nu}_c$ \& $\bm{\omega}_c$ are the camera's linear and angular velocities, respectively, and $\mathbf{y}$ is the output vector.  

\subsection{An Observer for a Class of Nonlinear Systems}
\label{sec:obs_nonlinear}

Several observer classes were studied in the automatic control community.
For the purpose of this work, we are interested in the observer proposed in \cite{spica2013}, for the class of systems described by
\begin{equation}
    \begin{cases}
        \dot{\mathbf{s}} = \mathbf{f}_m (\mathbf{s}, \bm{\omega}_c) + \bm{\Omega}^T(\mathbf{s},\bm{\nu}_c)\bm{\chi} \\
        \dot{\bm{\chi}} = \mathbf{f}_u(\mathbf{s},\bm{\chi},\mathbf{v}_c),
    \end{cases}
    \label{eq:sys_class}
\end{equation}
where $\mathbf{s} \in \mathbb{R}^m$ is the vector of the measurable components of the state, $\bm{\chi} \in \mathbb{R}^p$ the unknown components, and $f_m(.)$ \& $f_u(.)$ are sufficiently smooth w.r.t. their arguments. 
Let $[\hat{\mathbf{s}},\hat{\bm{\chi}}]^T$ be the estimated state, and $[\tilde{\mathbf{s}},\tilde{\bm{\chi}}]^T$ be the state estimation error, such that  $\tilde{\mathbf{s}} = \mathbf{s} - \hat{\mathbf{s}}$ and $\tilde{\bm{\chi}} = \bm{\chi} - \hat{\bm{\chi}}$.
Then, the observer given by 
\begin{equation}
    \label{eq:obs_class}
    \begin{cases}
        \dot{\hat{\mathbf{s}}} = \mathbf{f}_m (\mathbf{s}, \bm{\omega}_c) + \bm{\Omega}^T(\mathbf{s},\bm{\nu}_c)\hat{\bm{\chi}} + \mathbf{H} \tilde{\mathbf{s}}  \\
        \dot{\hat{\bm{\chi}}} = \mathbf{f}_u(\mathbf{s},\hat{\bm{\chi}},\mathbf{v}_c) + \alpha \bm{\Omega}(\mathbf{s},\bm{\nu}_c)  \tilde{\mathbf{s}}.
    \end{cases}
\end{equation}
with $\mathbf{H} \succ 0$, and $\alpha > 0$, recovers the system's state.

Given the system defined in \eqref{eq:ddyn}, \eqref{eq:hdyn}, \eqref{eq:depthdyn} \& \eqref{eq:y}, and the fact that only the moment vector $\mathbf{h}$ is measurable (computed directly from the image plane \cite{hartley2003}), it is possible to conclude that, for this system $\mathbf{s} \equiv \mathbf{h}$ and $\bm{\chi} \equiv [\mathbf{d}\ l]^T$. 
From inspection of the system in \eqref{eq:sys_class}, $\bm{\chi}$ should appear linearly in the dynamics of $\mathbf{s}$.
Since, in \eqref{eq:hdyn} the depth appears in the denominator, that condition is not verified.
However, given the change of coordinates proposed in \cite{mateus2018}
\begin{equation}
    \bm{\chi} = \frac{\mathbf{d}}{l},
    \label{eq:chi}
\end{equation}
the condition will be verified by replacing \eqref{eq:chi} in \eqref{eq:hdyn}.
The new moment vector dynamics are 
\begin{equation}
    \dot{\mathbf{h}} =  \bm{\omega}_c \times \mathbf{h} - \bm{\nu}_c^T\mathbf{h}( \bm{\chi} \times \mathbf{h})
    \label{eq:hdyn_new},
\end{equation}
and the dynamics of $\bm{\chi}$ are given by
\begin{equation}
    \dot{\bm{\chi}} = \bm{\omega}_c \times \bm{\chi} - \bm{\chi} \bm{\nu}_c^T(\bm{\chi} \times \mathbf{h}).
    \label{eq:chidyn}
\end{equation}
Now, the dynamical system given by \eqref{eq:chidyn}, and \eqref{eq:hdyn_new} belongs to the class of systems in \eqref{eq:sys_class}, with 
\begin{equation}
    \bm{\Omega} = - \bm{\nu}^T \mathbf{h} [\mathbf{h}]_\times,
    \label{eq:omega}
\end{equation}
where $[\mathbf{h}]_\times$ is a $3\times3$ skew-symmetric matrix that linearizes the cross product (such that $\bm{\chi} \times \mathbf{h} = - [\mathbf{h}]_\times\bm{\chi}$). 

The \emph{persistence of excitation condition} in \cite{deluca2008} gives an observability criterion for the observer in \eqref{eq:obs_class}. 
This means that the state estimation error approaches zero iff the matrix $\bm{\Omega}\bm{\Omega}^T$ is full rank. 
From \eqref{eq:omega}, this is not the case since the skew-symmetric matrix has rank $2$ by definition.  
Thus, convergence of the observer cannot be guaranteed for this dynamical system.
A solution is to couple the orthogonality of the \emph{Pl\"ucker Coordinates} given by \eqref{eq:ortho} with the state dynamics, yielding a Differential-Algebraic System.

In the next section, a method to introduce the constraint directly on the dynamics is proposed.
This method results in a dynamical system, which verifies the observability criterion of the observer \eqref{eq:obs_class}.
Hence, avoiding the need to design an observer for the DAS.

\section{From Euclidean to Spherical coordinates}
\label{sec:spherical}


In this paper we aim at representing lines with four degrees of freedom (minimum number), while keeping the advantages of the \emph{Pl\"ucker Coordinates}.
In a first step, the unit vector $\mathbf{h}$ is converted to spherical coordinates being, then, represented by two spherical angles.
Secondly, two unit vectors orthogonal to $\mathbf{h}$ and to each other are computed, allowing us to construct an orthonormal basis.
The vector $\bm{\chi}$ in \eqref{eq:chi} is projected into that basis, and its projection onto the axis $\mathbf{h}$ is zero due to \eqref{eq:ortho}.
The new system's state is identified, and its dynamics are computed.
It belongs to the class of systems defined in \eqref{eq:sys_class}, and satisfies the observability criterion.
Thus, an observer based on \eqref{eq:obs_class} is designed, which is presented in Sec.~\ref{sec:obs_active}.

\subsection{Change of Coordinates and Orthogonal Basis}
\label{sec:spherical_basis}

Let us define $\mathbf{h}$ using spherical coordinates
\begin{equation}
    \mathbf{h}_{S} = \begin{bmatrix} r \cos(\theta) \cos(\phi)\\
                    r \sin(\theta) \cos(\phi)\\
                    r \sin(\phi)
                \end{bmatrix}
    \label{eq:hsph},
\end{equation}
where $r$ is the length of $\mathbf{h}$ (one in this case), $-\pi \leq \theta \leq \pi$ is the azimuth, and $-\frac{\pi}{2} \leq \phi \leq \frac{\pi}{2}$ is the zenith angle.
Consider $\mathbf{h}_P$ to be orthogonal to $\mathbf{h}_S$, which can be defined by
\begin{equation}
    \mathbf{h}_P = \begin{bmatrix} r \cos(\theta) \sin(\phi)\\
                    r \sin(\theta) \sin(\phi)\\
                    - r \cos(\phi)
                \end{bmatrix}
    \label{eq:hpsph}.
\end{equation}
Since both $\mathbf{h}_S$, and $\mathbf{h}_P$ are unitary, let us define an orthonormal basis $\mathbf{A}$ by
\begin{equation}
    \mathbf{A} = \begin{bmatrix}
                    \mathbf{h}_S^T \\
                    \mathbf{h}_P^T\\
                    (\mathbf{h}_S \times \mathbf{h}_P)^T
                \end{bmatrix}
    \label{eq:obA},
\end{equation}
with $\mathbf{A}^T\mathbf{A} = \mathbf{I}$, where $\mathbf{I}$ is the identity matrix, and
\begin{equation}
    (\mathbf{h}_S \times \mathbf{h}_P) = \begin{bmatrix} -\sin(\theta) \\ \cos(\theta) \\ 0 \end{bmatrix}.
    \label{eq:hsxhp}
\end{equation}
Making use of \eqref{eq:ortho}, we define
\begin{equation}
    \bm{\eta} = \mathbf{A}\bm{\chi} = \begin{bmatrix} 0 \\ \eta_1 \\ \eta_2
                                      \end{bmatrix}
    \label{eq:eta}.
\end{equation}
Our new state variables are $\theta$, $\phi$, $\eta_1$, and $\eta_2$. In order to define a new dynamical system we require the dynamics of these variables, which are presented in the next subsection.

\subsection{Proposed Dynamics}
\label{sec:proposed_dyn}

Now, using the change of coordinates proposed above, in this subsection the new system dynamics are computed.
Given that $\mathbf{A}$ it is an orthogonal matrix is easy to conclude that
\begin{equation}
    \bm{\chi} = \mathbf{A}^T\bm{\eta}
    \label{eq:var_change_chi}.
\end{equation}
Replacing \eqref{eq:var_change_chi} and \eqref{eq:hsph} in \eqref{eq:hdyn_new}, the moment's dynamics become
\begin{equation}
    \dot{\mathbf{h}} =  \bm{\omega}_c \times \mathbf{h}_S + \bm{\Omega}^T\mathbf{A}^T\bm{\eta}.
    \label{eq:h_basis_change}
\end{equation}
In addition, taking the time derivative of \eqref{eq:hsph}:
\begin{equation}
    \dot{\mathbf{h}}_{S} = 
     \begin{bmatrix} \mathbf{h}_S  & -\mathbf{h}_P & (\mathbf{h}_S \times \mathbf{h}_P) \cos(\phi) \end{bmatrix}
    \begin{bmatrix} \dot{r} \\ \dot{\phi} \\ \dot{\theta} \end{bmatrix}
    \label{eq:hsph_dyn},
\end{equation}
equaling \eqref{eq:hsph_dyn} \& \eqref{eq:h_basis_change}, and solving for $\dot{r}$, $\dot{\theta}$, and $\dot{\phi}$ we get
\begin{align}
    \dot{r} = &~ 0 \label{eq:dr}\\ 
    \dot{\theta} = &~ \frac{-\bm{\omega}_c^T  \mathbf{h}_P  + \bm{\nu}_c^T\mathbf{h}_S\eta_1}{\cos(\phi)} \label{eq:dt}\\ 
    \dot{\phi} = &~ -\bm{\omega}_c^T (\mathbf{h}_S \times \mathbf{h}_P) + \bm{\nu}_c^T\mathbf{h}_S \eta_2  .
    \label{eq:dp}
\end{align}
Then, from \eqref{eq:sys_class} we define
\begin{equation}
    \bm{\Omega}_S = \bm{\nu}_c^T\mathbf{h}_S \begin{bmatrix} \frac{1}{\cos(\phi)} & 0 \\ 0 & 1 \end{bmatrix}
    \label{eq:omega_s}
\end{equation}
which is diagonal, thus it is full rank as long as $\bm{\nu}_c^T\mathbf{h}_S \neq 0$ ($\bm{\Omega}_S = 0$).
Meaning, that it is possible to observe the system's state, provided that the linear velocity vector of the camera does not belong to the line's interpretation plane. 
Notice that, when the moment vector is coincident with any of the Cartesian axis, one should pay special attention to the definition of the spherical angles, since  for $\phi = \pm\pi/2$, $\bm{\Omega}_S$ will not be defined.

In order to have the remaining state dynamics we have to define $\dot{\eta}_1$, and $\dot{\eta}_2$. Let us take the time derivative of \eqref{eq:eta}:
\begin{equation}
    \dot{\bm{\eta}} = \dot{\mathbf{A}}\bm{\chi} + \mathbf{A}\dot{\bm{\chi}},
    \label{eq:deta}
\end{equation}
where
\begin{equation}
    \dot{\mathbf{A}} = \begin{bmatrix} \dot{\mathbf{h}}_S^T \\ \dot{\mathbf{h}}_P^T \\ (\dot{\mathbf{h}_S \times \mathbf{h}_P})^T
    \end{bmatrix}.
    \label{eq:dA}
\end{equation}
Finally replacing \eqref{eq:dA}, \eqref{eq:chidyn}, and \eqref{eq:var_change_chi} in \eqref{eq:deta} yields
\begin{multline}
    \dot{\eta}_1 = -\bm{\omega}_c^T \left( \mathbf{h}_P\tan(\phi) + \mathbf{h}_S \right)\eta_2 + \\
   + \bm{\nu}_c^T \left(  \left(\mathbf{h}_S\tan(\phi) - \mathbf{h}_P \right)\eta_1\eta_2 + (\mathbf{h}_S\times\mathbf{h}_P)\eta_1^2\right) =\\
   = f_{\eta_1}(\theta,\phi,\eta_1,\eta_2, \bm{\nu}_c, \bm{\omega}_c)
   \label{eq:eta1_dyn}
\end{multline}
and
\begin{multline}
    \dot{\eta}_2 = \bm{\omega}_c^T\left( \mathbf{h}_P \tan(\phi) + \mathbf{h}_S \right)\eta_1 + \\
    + \bm{\nu}_c^T \left( (\mathbf{h}_S\times\mathbf{h}_P)\eta_1\eta_2 - \mathbf{h}_S\tan(\phi)\eta_1^2 - \mathbf{h}_P\eta_2^2 \right) = \\
    = f_{\eta_2}(\theta,\phi,\eta_1,\eta_2, \bm{\nu}_c,\bm{\omega}_c)
    \label{eq:eta2_dyn}.
\end{multline}

In the new coordinates, the dynamical system's state is given by $\theta$, $\phi$, $\eta_1$, and $\eta_2$.
Its inputs are the camera velocities (as before), the outputs is given as
\begin{equation}
    \mathbf{y} = \begin{bmatrix} \theta \\ \phi \end{bmatrix},
    \label{eq:y_new}
\end{equation}
and the dynamics are given by \eqref{eq:dt}, \eqref{eq:dp}, \eqref{eq:eta1_dyn}, and \eqref{eq:eta2_dyn}. In the next section an observer for this system is presented, with a method to optimize the convergence behavior.

\section{Observer Design and Active Estimation}
\label{sec:obs_active}

\begin{figure*}
    \vspace{-.45cm}
    \centering
    \subfloat[Real and estimated state evolution.]{
        \includegraphics[width=0.25\textwidth]{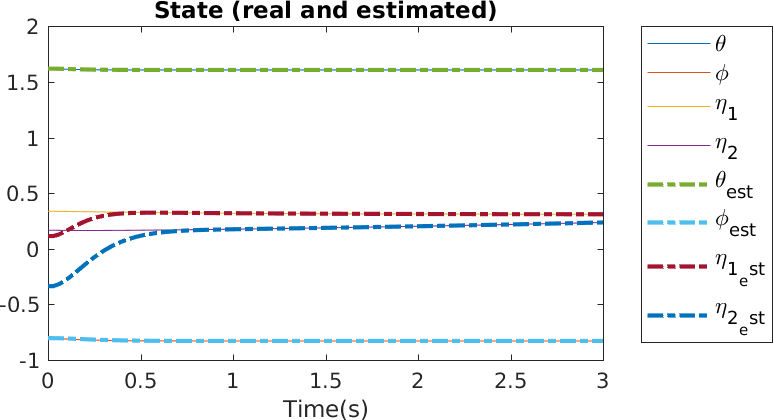}
        \label{fig:sim3a}
    }
    \hfill
    \subfloat[Camera's Linear and Angular Velocities.]{
        \includegraphics[width=0.25\textwidth]{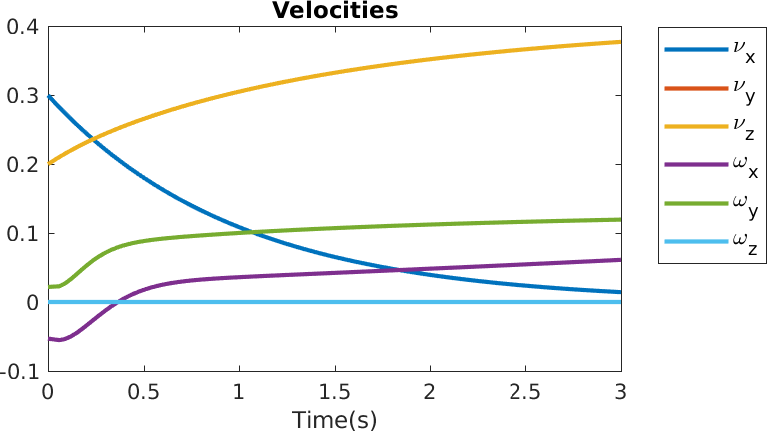}
        \label{fig:sim3c}
    }
    \hfill
    \subfloat[Evolution of the Eigenvalues of the matrix $\bm{\Omega}\bm{\Omega}^T$]{
        \includegraphics[width=0.2\textwidth]{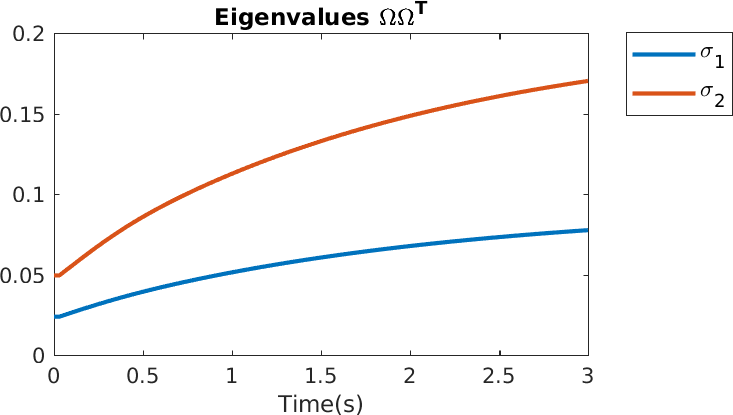}
        \label{fig:sim3d}
    }
    \hfill
    \subfloat[3D plot of the simulated world]{
        \includegraphics[width=0.185\textwidth]{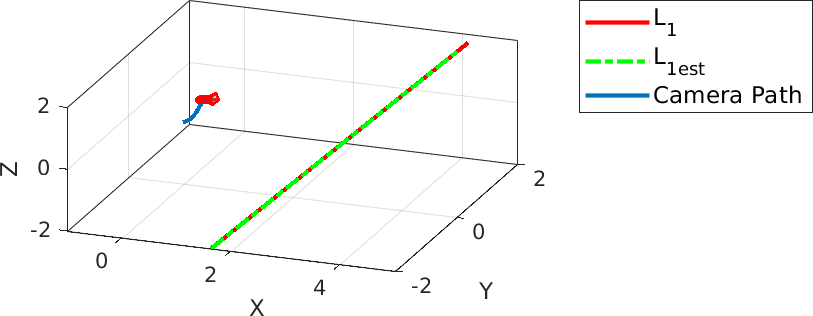}
        \label{fig:sim3e}
    }
    \caption{Simulation results for a single line. The actual and the estimated state are presented in the first left plot. The velocities of the camera are presented in the second plot. The eigenvalues of the matrix $\bm{\Omega}_S\bm{\Omega}_S^T$ are presented in the third plot. Finally, a 3D plot of the environment with the actual \& estimated lines, the camera and its path are presented in the right plot.}
    \label{fig:sim3}
\end{figure*}

The new dynamical system presented in Sec.~\ref{sec:spherical} is defined by \eqref{eq:dt}, \eqref{eq:dp}, \eqref{eq:eta1_dyn}, \eqref{eq:eta2_dyn} \& \eqref{eq:y_new}, and belongs to the class of systems in  \eqref{eq:sys_class}.
Thus, it is possible to design an observer to estimate the system's state based on \eqref{eq:obs_class}.
Hence, following the design procedure in \cite{spica2013} the observer
\begin{align}
\begin{split}
    \begin{bmatrix} \dot{\hat{\theta}} \\ \dot{\hat{\phi}}\end{bmatrix} = & -\bm{\omega}_c^T \begin{bmatrix} \frac{\mathbf{h}_P}{\cos(\phi)} \\ (\mathbf{h}_S \times \mathbf{h}_P) \end{bmatrix}  + \bm{\Omega}_S^T  \begin{bmatrix} \hat{\eta_1} \\ \hat{\eta_2} \end{bmatrix} + \mathbf{H}_S \begin{bmatrix} \tilde{\theta} \\ \tilde{\phi} \end{bmatrix}\\
    \begin{bmatrix} \dot{\hat{\eta}}_1\\ \dot{\hat{\eta}}_2\end{bmatrix} = & \begin{bmatrix} f_{\eta_1}(\theta,\phi,\hat{\eta_1},\hat{\eta_2}, \bm{\nu}_c, \bm{\omega}_c) \\ f_{\eta_2}(\theta,\phi,\hat{\eta_1},\hat{\eta_2}, \bm{\nu}_c,\bm{\omega}_c) \end{bmatrix} + \alpha\bm{\Omega_S} \begin{bmatrix} \tilde{\theta} \\ \tilde{\phi} \end{bmatrix}
\end{split},
    \label{eq:spherical_observer}
\end{align}
is proposed. 
Where $\mathbf{H}_S \succ 0$, $\tilde{\theta} = \theta - \hat{\theta}$, and $\tilde{\phi} = \phi - \hat{\phi}$.

Let $\mathbf{U}\bm{\Sigma}\mathbf{V} = \bm{\Omega}_S$ be the singular value decomposition of matrix $\bm{\Omega}_S$, where $\bm{\Sigma} = \text{diag}(\{\sigma_i\})$, $i = 1,2$, with $\sigma_i$ being the singular values from lowest to highest. 
Then, $\mathbf{H}_S \in \mathbb{R}^{2 \times 2}$ may be chosen as  
\begin{equation}
    \mathbf{H}_S = \mathbf{V} \mathbf{D} \mathbf{V}^T,
\end{equation}
where $\mathbf{D} \in \mathbb{R}^{2\times 2}$ is a function of the singular values of $\bm{\Omega}_S$.
Following \cite{spica2013}, the former is defined as $\mathbf{D} = \text{diag}(\{c_i\})$, with $c_i > 0$, and $c_i = 2\sqrt{\alpha}\sigma_i$, for $i = 1,2$.
This choice prevents oscillatory modes, thus trying to achieve a critically damped transient behavior.

As stated previously the \emph{persistence of excitation condition} is an observability criterion for the observer \eqref{eq:spherical_observer}. 
This criterion implies that the matrix $\bm{\Omega}_S\bm{\Omega}_S^T$ be full rank, which from \eqref{eq:omega_s} is verified. 
It can be proven that the convergence rate depends on its smallest eigenvalue (i.e. $\sigma_1^2$).
Since $\bm{\Omega}_S$ is diagonal the eigenvalues are given by the diagonal entries, which depend on the measurements $\mathbf{h}_S$, and on the camera's linear velocity.
There are two ways the convergence behavior can be acted upon.
The first consists in increasing the gain $\alpha$, and the second in choosing the camera's linear velocity to maximize $\sigma_1^2$.
The former solution is more straightforward, but it may result in higher sensitivity to noise. 
Thus, in this work the latter solution is used.

Let us compute the total time derivative of the eigenvalues of matrix $\bm{\Omega}_S\bm{\Omega}_S^T$.
Using the results in \cite{spica2013}, we conclude that 
\begin{equation}
   \dot{\sigma}_i^2 = \mathbf{J}_{\bm{\nu}_c,i} \dot{\bm{\nu}}_c + \mathbf{J}_{\theta,\phi,i} \begin{bmatrix} \dot{\theta} \\ \dot{\phi} \end{bmatrix} \quad \text{with} \; i = 1,2,
   \label{eq:dsigma}
\end{equation}
where matrices $\mathbf{J}_{\bm{\nu}_c,i}$  $\mathbf{J}_{\mathbf{s},i}$ are the Jacobian matrices of the eigenvalues of $\bm{\Omega}_S\bm{\Omega}_S^T$ w.r.t. the linear velocity, and $[\theta, \phi]^T$ respectively.
The Jacobian for the linear velocity is given by
\begin{equation}
    \mathbf{J}_{\bm{\nu}_c} = \bm{\nu}_c^T\mathbf{h}_S  \begin{bmatrix} \frac{2\cos(\theta)}{\cos(\phi)} & \frac{2\sin(\theta)}{\cos(\phi)} & \frac{2\sin(\phi)}{\cos(\phi)^2} \\  2\cos(\theta)\cos(\phi) & 2\sin(\theta)\cos(\phi) & 2\sin(\phi) \end{bmatrix}.
\end{equation}
A differential inversion technique can be used to regulate the eigenvalues, by acting on $\dot{\bm{\nu}}_c$.
However, this does not account for the effect of the second term on the right-hand side of \eqref{eq:dsigma}.
A solution is to enforce $ [\dot{\theta} , \dot{\phi}]^T \simeq \mathbf{0}$.

As stated above the convergence rate depends on $\sigma_1^2$, which is higher the higher the norm of the linear velocity of the camera, resulting in a faster convergence. 
However, increasing $||\bm{\nu}_c||$, implies an increase in the control effort, which may have practical problems (e.g. damage to the motors).
Thus, a control law is designed to keep $||\bm{\nu}_c||$ relatively low, and at the same time driving the eigenvalues ($\sigma_i^2$) to a desired value. The control law used is given by 
\begin{equation}
    \dot{\bm{\nu}}_c = k_1 \mathbf{J}_{\bm{\nu}_c}^{\dagger} (\bm{\sigma}_{des}^2 - \bm{\sigma}^2) + k_2 \left( \mathbf{I}_2 - \mathbf{J}_{\bm{\nu}_c}^{\dagger}\mathbf{J}_{\bm{\nu}_c} \right)\bm{\nu}_c,
    \label{eq:control_law}
\end{equation}
which as been proposed in \cite{spica2013}. Where $\mathbf{I}_2 \in \mathbb{R}^{2\times2}$ is an identity matrix, $k_1 > 0$ and $k_2 > 0$ are constant gains, $\mathbf{J}_{\bm{\nu}_c}^{\dagger}$ is the Moore-Penrose pseudo-inverse of the Jacobian matrix, $\bm{\sigma}^2 = [\sigma_1^2,\sigma_2^2]^T \in \mathbb{R}^2$, and $\bm{\sigma}_{des}^2$ are the desired eigenvalues.

The effect of $\dot{\theta}$ and $\dot{\phi}$ in the dynamics of the eigenvalues is compensated, using the camera's angular velocity.
It is computed by setting \eqref{eq:dt} and \eqref{eq:dp} to zero and solving for $\bm{\omega}_c$ yielding
\begin{equation}
    \bm{\omega}_c = \frac{(\bm{\nu}_c^T\mathbf{h}_S)}{\sin(\phi)} \begin{bmatrix} \eta_1\cos(\theta) - \eta_2\sin(\theta)\sin(\phi) \\ \eta_1\sin(\theta) + \eta_2\cos(\theta)\sin(\phi) \\ 0 \end{bmatrix}.
    \label{eq:wc}
\end{equation}
The next section presents the results for the proposed method both in simulation and in a real environment.

\section{Results}
\label{sec:results}

In this section we validate the observer and active estimation scheme proposed in Sec.~\ref{sec:obs_active} for the system presented in Sec.~\ref{sec:spherical}.
First, with an experiment in simulation, then, the method is evaluated in a real robot \cite{messias2014}.

\begin{figure*}
    \vspace{-.4cm}
    \centering
    \subfloat[Estimated state evolution.]{
        \includegraphics[width=0.22\textwidth]{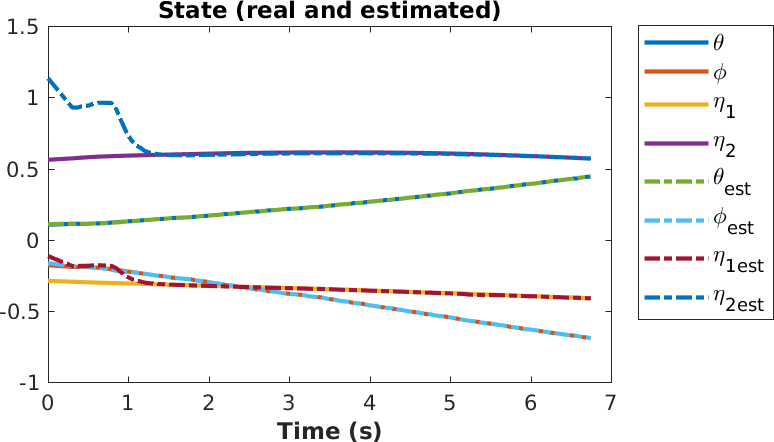}
        \label{fig:real_linea}
    }
    \hfill
    \subfloat[State estimation error over time.]{
        \includegraphics[width=0.22\textwidth]{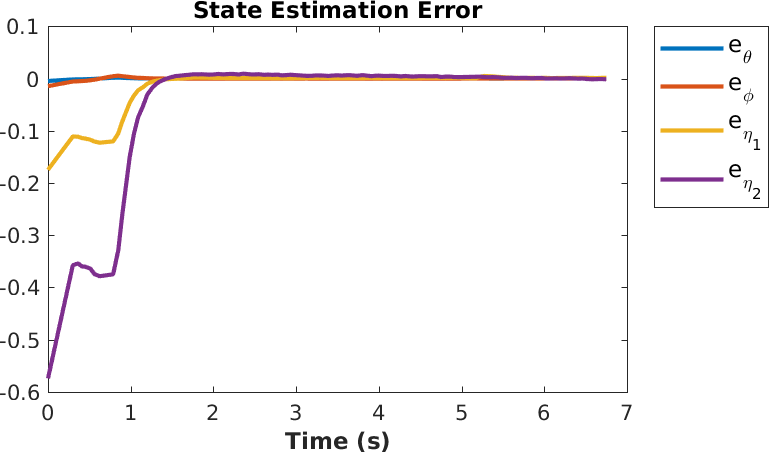}
        \label{fig:real_lineb}
    }
    \hfill
    \subfloat[Camera's Linear and Angular Velocities.]{
        \includegraphics[width=0.22\textwidth]{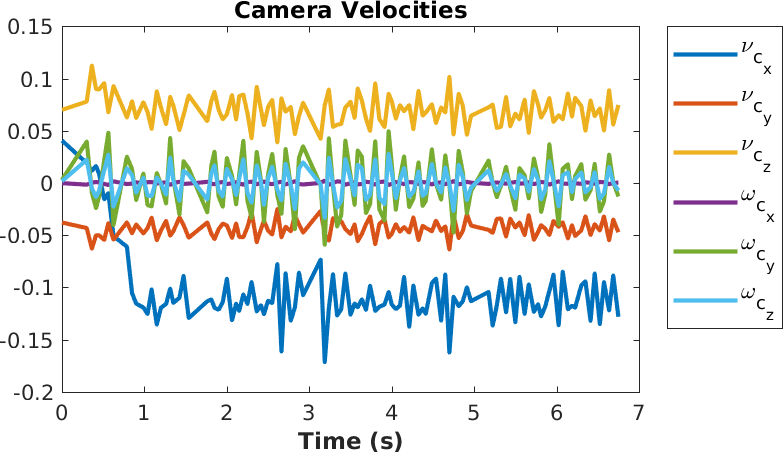}
        \label{fig:real_linec}
    }
    \hfill
    \subfloat[Evolution of the Eigenvalues of the matrix $\bm{\Omega}\bm{\Omega}^T$]{
        \includegraphics[width=0.22\textwidth]{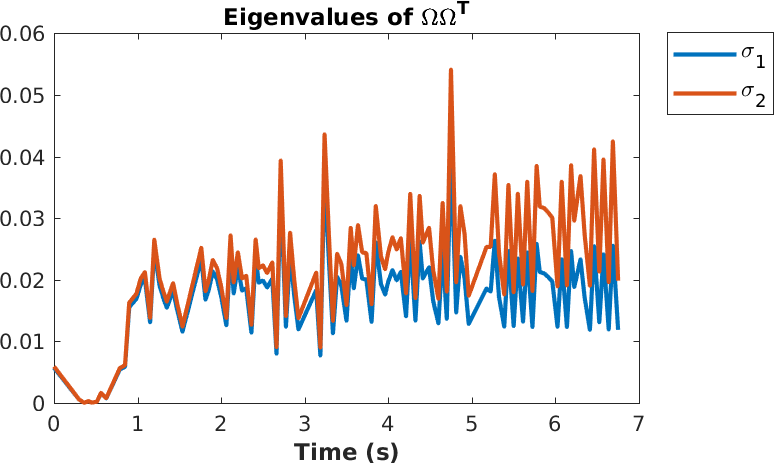}
        \label{fig:real_lined}
    }\\
    \subfloat[3D plot of the world]{
        \includegraphics[width=0.35\textwidth]{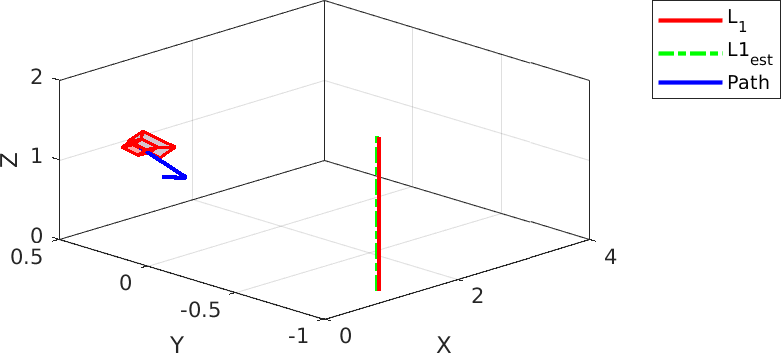}
        \label{fig:real_linee}
    }
    \;\;
    \subfloat[Robotic Platform.]{
        \includegraphics[height=0.152\textheight]{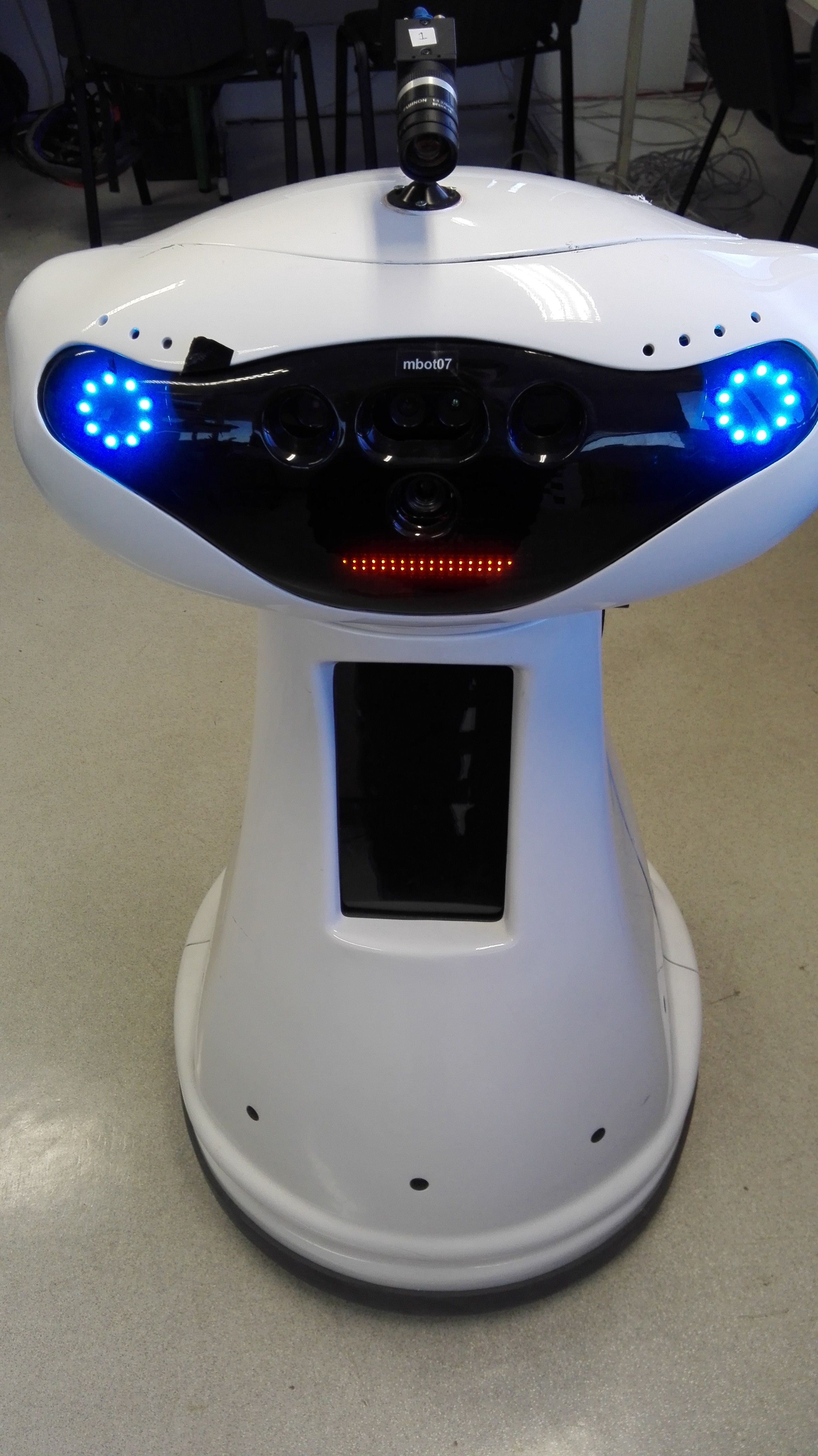}
        \label{fig:mbot}
    }
    \;\;
    \subfloat[Camera Image with tracked lines and points for pose estimation.]{
        \includegraphics[height=0.152\textheight]{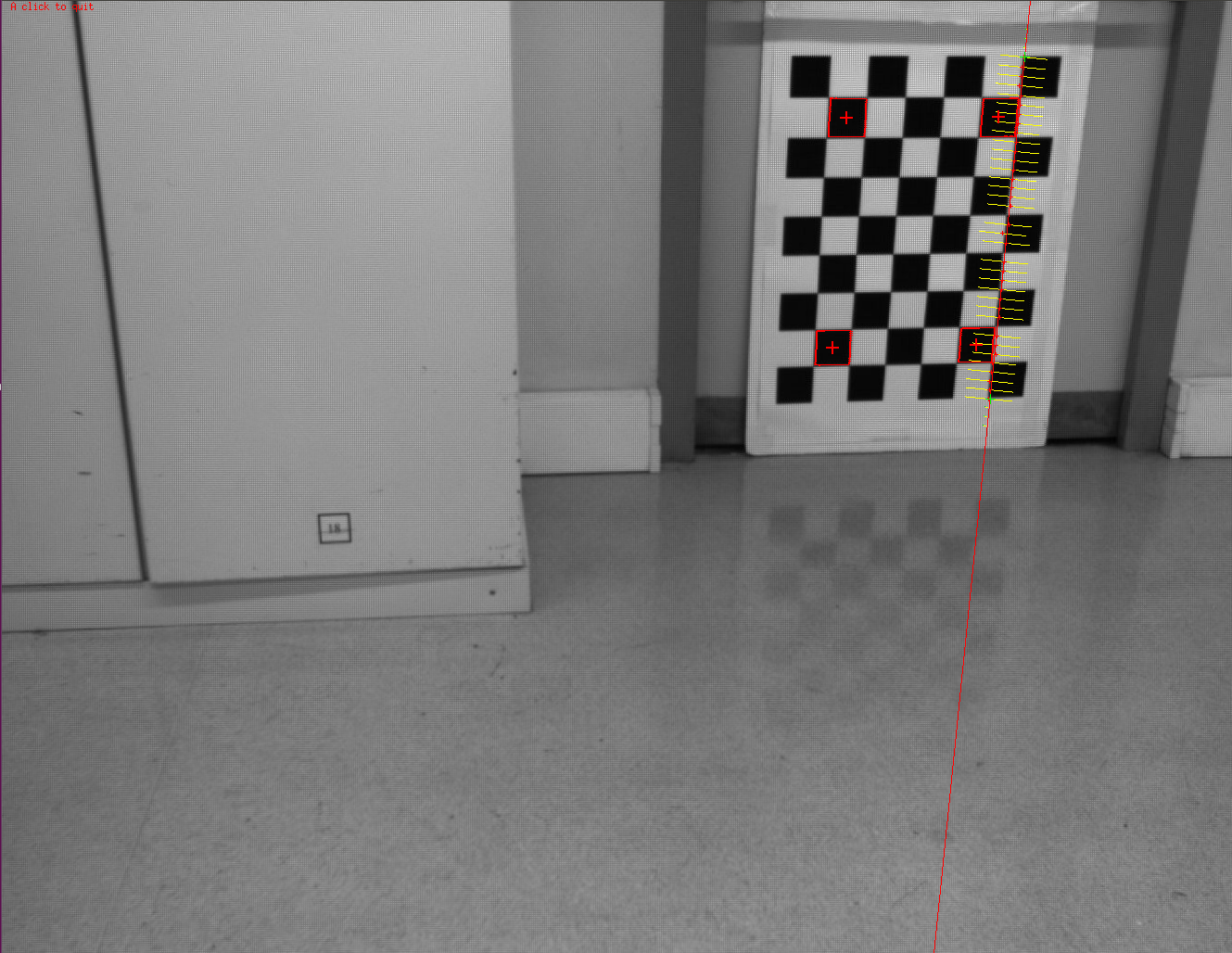}
        \label{fig:line_track}
    }%
    \caption{Experimental results of the active state estimation with a real robot and camera. The real and estimated state are presented in the top left plot. The state estimation error is presented in the second top plot. The velocities in the camera reference frame are presented in the third top plot. The eigenvalues of the matrix $\bm{\Omega}\bm{\Omega}^T$ are presented in the fourth top plot. A 3D plot of the real \& estimated lines in the world reference frame, the camera and its path are plotted in the bottom left. In the middle bottom is the robotic platform used, with a Pointgrey Flea3 USB3 camera on top. The four points, of the chessboard, used for pose estimation and the line tracked is shown on the bottom right. }
    \label{fig:real_line}
\end{figure*}

\subsection{Simulation Results}
\label{sec:simulation}

The observer, in \eqref{eq:spherical_observer}, and the active estimation scheme comprised of the control laws in \eqref{eq:control_law} \& \eqref{eq:wc} were simulated in MATLAB.
A perspective camera was used as the imaging device.
The intrinsic parameters matrix (for more details see \cite{hartley2003}) is defined as $\mathbf{I}_3 \in \mathbb{R}^{3\times3}$. 
All six degrees-of-freedom (DoF) are assumed to be controllable. 
The actual coordinates of the line are obtained by generating a random point within a cube with $4[\text{m}]$ side in front of the camera, then a unit direction is randomly selected.
The moment vector is computed using \eqref{eq:h_def}, then, the change of variables in \eqref{eq:chi} is applied and, the moment vector is transformed to spherical coordinates.
Matrix $\mathbf{A}$ is computed, and $\eta_1$ and $\eta_2$ are obtained with \eqref{eq:eta}.
The initial estimate  $(\hat{\eta}_1,\hat{\eta}_2)$ is also generated randomly, $\hat{\mathbf{h}}_S$ is initialized with its true value, since it is available from the measurements.

Fig.~\ref{fig:sim3} presents results for a simulation, with the following gains $k_1 , k_2 = 1$, $\alpha = 2000$, and $\bm{\sigma}_{des}^2 = [0.08,0.18]^T$ (see \eqref{eq:control_law} and \eqref{eq:spherical_observer}).
Fig.~\ref{fig:sim3}\subref{fig:sim3a} shows the evaluation of the state's evolution for both the actual system and its estimate. 
The objective $[\dot{\theta} , \dot{\phi}]^T \simeq \mathbf{0}$ was achieved, and convergence was achieved in less than 1 second.
Fig.~\ref{fig:sim3}\subref{fig:sim3c} shows the velocities. 
Fig.~\ref{fig:sim3}\subref{fig:sim3d} presents the variation of the eigenvalues over time.
Finally, Fig.~\ref{fig:sim3}\subref{fig:sim3e} presents a plot with the actual and estimated lines. Besides it presents the camera and its path in the world coordinates (assumed to be the camera's initial pose).
The total error in the \emph{Pl\"ucker Coordinates} is $||\mathbfcal{L} - \mathbfcal{L}_{est}|| = 6.37\times10^{-4}$, where $\mathbfcal{L}$ and $\mathbfcal{L}_{est}$ are the actual and the estimated coordinates of the line, respectively. 

\subsection{Experimental Results}
\label{sec:exp_results}


This section presents the experimental results using a real robotic platform, MBOT \cite{messias2014} (see Fig.~\ref{fig:real_line}\subref{fig:mbot}).
The robot has an omnidirectional base, having therefore 3 DoF (2 linear, and 1 angular).
The observer and the active estimation method were implemented using the Robot Operating System (\emph{ROS}) \cite{quigley2009}. 
In the experiment, the moving-edges tracker \cite{marchand2005b}, available in \emph{ViSP} \cite{marchand2005}, was used to track lines in the image as shown in Fig.~\ref{fig:real_line}\subref{fig:line_track}.
The images were provided by a Pointgrey Flea3 USB3\footnote{https://www.ptgrey.com/flea3-usb3-vision-cameras}, which was mounted on the top of the robot.

In this experiment, we attempt to replicate the results in Sec.~\ref{sec:simulation}, in order to show that the performance is kept in a real setup.
For this purpose, the estimation of a line in a chessboard was tested. 
A chessboard was used to define the line, since one can easily and robustly compute the pose of the chessboard w.r.t. to the camera's reference frame, which will be used as ground-truth.
There are several algorithms to compute the chessboard's pose with known dimensions.
In this work the \emph{POSIT} algorithm, presented in \cite{oberkampf1996}, was used.
The camera was calibrated extrinsically a priori, i.e., the transformation between the robot and the camera's coordinate system was computed beforehand. 

The results of this experiment are presented in Fig.~\ref{fig:real_line}, with control gains (see \eqref{eq:control_law}) $k_1 = k_2 = 1$, $\alpha = 2000$, and the desired eigenvalues are $\bm{\sigma}_{des} = [0.2,0.2]^T$. 
Notice that the eigenvalues are set to the same value, since the line to estimate is vertical,
thus the $x$ coordinate of $\mathbf{h}$ (defined in \eqref{eq:h_def}) is close to $1$. 
Hence, the angle $\phi \simeq 0$, and $\cos(\phi) \simeq 1$, which from \eqref{eq:omega_s} yields a diagonal matrix, whose entries are almost identical.
The real and the estimated states are presented in Fig.~\ref{fig:real_line}\subref{fig:real_linea}.
Since the platform is a mobile robot, it can only move in a plane. 
This means that it only has three DoF, two linear from moving in both directions, and 1 angular from rotating about the normal to the plane. 
From \eqref{eq:wc}, it can be concluded that to achieve $\dot{\mathbf{h}} \simeq \mathbf{0}$, the robot should have two DoF in rotation. Thus, this goal was not achieved.
The state estimation error is presented in Fig.~\ref{fig:real_line}\subref{fig:real_lineb}.
The observer convergences in about 2 seconds. 
The velocities in the camera reference frame are presented in Fig.~\ref{fig:real_line}\subref{fig:real_linec}.
The velocity readings are noisy, because they are retrieved from the robot's odometry.
Fig.~\ref{fig:real_line}\subref{fig:real_lined} shows the variation of the the eigenvalues of the matrix $\bm{\Omega}\bm{\Omega}^T$. 
Notice that, despite the noisy velocity readings, the eigenvalues approached the desired value.
Finally, in Fig.~\ref{fig:real_line}\subref{fig:real_linee} presents the real and estimated lines in the world coordinates (assumed to be the camera's initial pose). Besides, it presents the camera pose in the end of the experiment, and its path throughout.
The error of the \emph{Pl\"ukcer Coordinates} estimated is $||\mathbfcal{L} - \mathbfcal{L}_{est}|| = 2\times10^{-3}$.

\section{Conclusions}
\label{sec:conclusions}

This work has proposed a method for active estimation of 3D straight lines with a monocular camera.
This method is based on two ideas.
The first is to represent the line's moment vector, which has unit norm, in spherical coordinates, thus reducing the number of variables needed to represent the vector.
The second is to use the unitary moment vector, and compute two unit vectors orthogonal to it and to each other. 
This yields an orthonormal matrix, which when pre-multiplied to the direction vector will result in a vector in $\mathbb{R}^3$ with a zero entry.
The dynamics in the new coordinates, spherical angles that define the moment vector and the projection of the direction vector to the new orthonormal basis are then computed.
An observer for this system is proposed based on a recent framework, followed by a control law that explores the camera's linear velocity to optimize the converge behavior.
The approach was validate both in simulation and with a real robotic platform.
Future work consists in adding additional constraints to the system, e.g. coplanarity constraints.

\bibliographystyle{IEEEtran}
\bibliography{active_sfm}

\end{document}